\def\year{2022}\relax
\documentclass[letterpaper]{article} % DO NOT CHANGE THIS
\usepackage{aaai22}  % DO NOT CHANGE THIS
\usepackage{times}  % DO NOT CHANGE THIS
\usepackage{helvet}  % DO NOT CHANGE THIS
\usepackage{courier}  % DO NOT CHANGE THIS
\usepackage[hyphens]{url}  % DO NOT CHANGE THIS
\usepackage{graphicx} % DO NOT CHANGE THIS
\usepackage{natbib}  % DO NOT CHANGE THIS AND DO NOT ADD ANY OPTIONS TO IT
\usepackage{caption} % DO NOT CHANGE THIS AND DO NOT ADD ANY OPTIONS TO IT
\DeclareCaptionStyle{ruled}{labelfont=normalfont,labelsep=colon,strut=off} % DO NOT CHANGE THIS
\usepackage{algorithm}
\usepackage{newfloat}
\usepackage{listings}

\usepackage{epsfig}
\usepackage{amsmath}
\usepackage{amssymb}
\usepackage{caption}
\usepackage{adjustbox}
\usepackage{wrapfig}
\usepackage{multirow}
\usepackage{algpseudocode}
\usepackage{arydshln}

\usepackage{subfigure}

\usepackage{color}
\usepackage{rotating}
\floatstyle{ruled}
\newfloat{listing}{tb}{lst}{}
\floatname{listing}{Listing}

\newcommand{\eg}{e.g.}
\newcommand{\ie}{i.e.}
\newcommand{\etal}{et~al.}
\vspace{-1.90em}
\title{FICGAN: Facial Identity Controllable GAN for De-identification}
\vspace{-0.90mm}
\author{Yonghyun Jeong\textsuperscript{\rm 1}, Jooyoung Choi\textsuperscript{\rm 2}, Sungwon Kim\textsuperscript{\rm 2}, Youngmin Ro\textsuperscript{\rm 1}, \\
Tae-Hyun Oh\textsuperscript{\rm 3}, Doyeon Kim\textsuperscript{\rm 1}, Heonseok Ha\textsuperscript{\rm 2}, Sungroh Yoon\textsuperscript{\rm 2}\\

}
\vspace{-0.90mm}
\affiliations {
    \textsuperscript{\rm 1}Samsung SDS,
    \textsuperscript{\rm 2}Seoul National University,
    \textsuperscript{\rm 3}POSTECH
} 

\usepackage{bibentry}

\begin{document}
% \maketitle
\twocolumn[{%
\renewcommand\twocolumn[1][]{#1}%
\maketitle
\begin{center}
    \centering
    \vspace{-17mm}
    \captionsetup{type=figure}
    \includegraphics[width=0.97\linewidth]{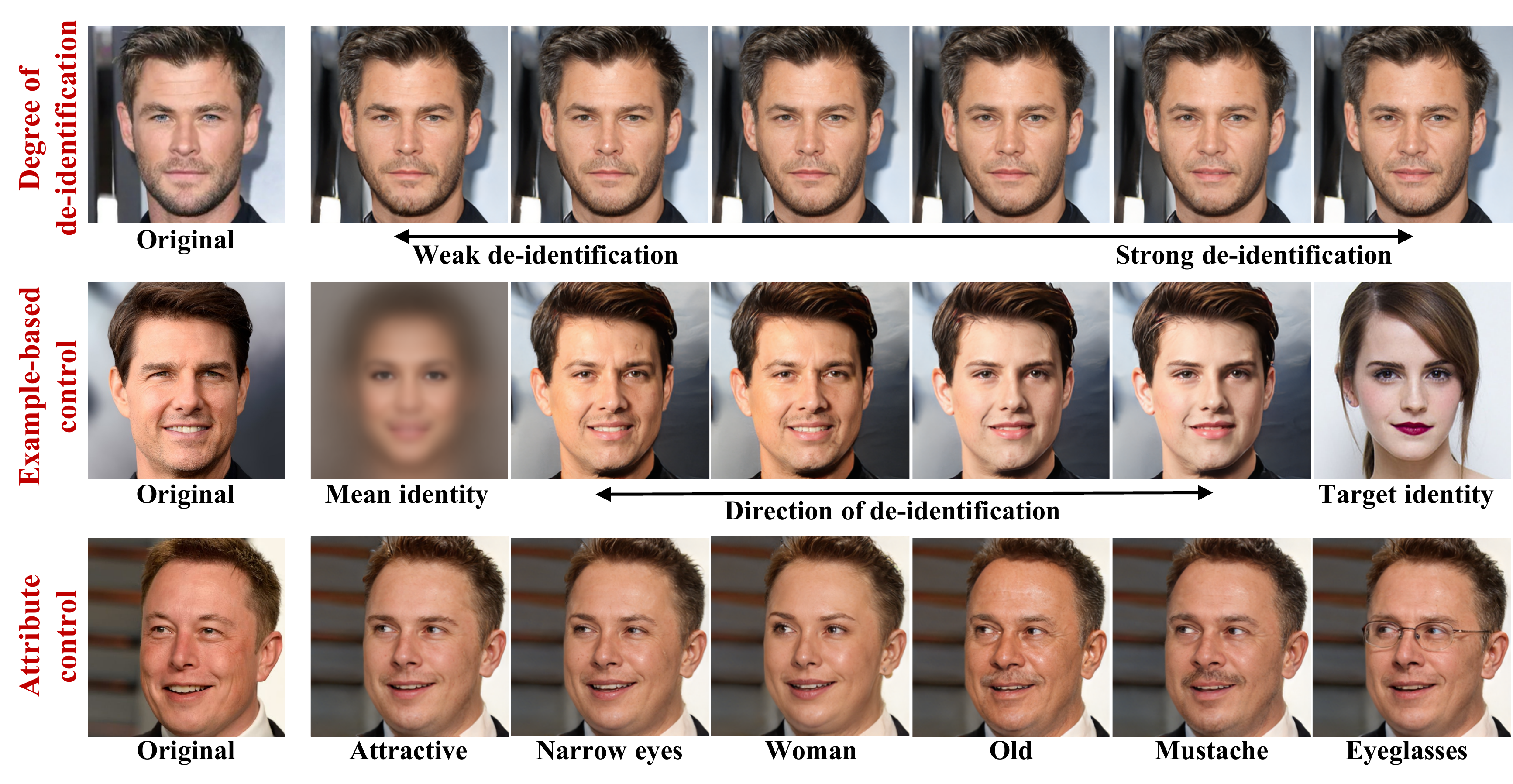}
    \vspace{-1.30mm}
    \captionof{figure}{\textbf{High-fidelity and controllable de-identification. } 
    Our method is able to manipulate in different levels of granularity. (Top row) Controlling the degree of de-identification.
    (Middle row) Example-based manipulation with preserving secure $k$-anonymity.
    (Bottom row) Attribute-wise manipulation with preserving $k$-anonymity.
    These examples show that our method strongly de-identifies faces under control, while the original pose, expression, shadow and illumination are well-preserved.}
    \label{fig:teaser}
    % \vspace{-0.3mm}
\end{center}%
}]

% %%%%%%%%% ABSTRACT
% \vspace{-30.90mm}
\begin{abstract}
\vspace{-1.5mm}
In this work, we present Facial Identity Controllable GAN (FICGAN) for not only generating high-quality de-identified face images with ensured privacy protection, but also detailed controllability on attribute preservation for enhanced data utility. We tackle the less-explored yet desired functionality in face de-identification based on the two factors. First, we focus on the challenging issue to obtain a high level of privacy protection in the de-identification task while uncompromising the image quality. Second, we analyze the facial attributes related to identity and non-identity and explore the trade-off between the degree of face de-identification and preservation of the source attributes for enhanced data utility. Based on the analysis, we develop Facial Identity Controllable GAN (FICGAN), an autoencoder-based conditional generative model that learns to disentangle the identity attributes from non-identity attributes on a face image. By applying the manifold $k$-same algorithm to satisfy $k$-anonymity for strengthened security, our method achieves enhanced privacy protection in de-identified face images. Numerous experiments demonstrate that our model outperforms others in various scenarios of face de-identification.
\end{abstract}
% \begin{figure*}[t]
%     \centering
%     %\captionsetup{type=figure}
%     \includegraphics[width=0.97\linewidth]{teaser_finalA.pdf}
%     \vspace{-0.90mm}
%     \caption{\textbf{High-fidelity and controllable de-identification.}
%     Our method is able to manipulate in different levels of granularity. (Top row) Controlling the degree of de-identification.
%     (Middle row) Example-based manipulation with preserving secure $k$-anonymity.
%     (Bottom row) Attribute-wise manipulation with preserving $k$-anonymity.
%     These examples show that our method strongly de-identifies faces under control, while the original pose, expression, shadow and illumination are well-preserved.}
%     \label{fig:teaser}
%     \vspace{-1.90mm}
% \end{figure*}

%%%%%%%%% BODY TEXT
\vspace{-3.0mm}
\section{Introduction}\vspace{-3.0mm}
Face de-identification refers to obscuring the identity of a person's face by manipulating the identity dominant attributes such as nose, eyes, eyebrows, and mouth, whereas the identity invariant attributes are preserved, such as pose, expression, shadow, and illumination.
The face is one of the most sensitive elements compare to other generic objects due to its personal identity information which is directly related to privacy.
Therefore, face de-identification is an important research topic in social aspects of security and privacy, and has been utilized to anonymize the faces that appeared in media interviews, the street-views on video surveillance contents, and medical research data~\cite{averaging, surveillance}.
The conventional methods, such as the black-bar masking, blurring, and pixelization, are simple but aggressive to remove the facial information, resulting in reduced data utility and unsuitable image quality~\cite{deid_overview,deepfake_de_id}.

Recently, deep learning methods have been employed for the task of face de-identification.
Thanks to the advances in Generative Adversarial Networks (GAN)~\cite{goodfellow2014generative}, simple methods such as face swap~\cite{faceswap} can be used to generate realistic outputs~\cite{faceshifter,id_invert}; 
however, since it employs the target person's face swapped onto the source image, it can be still identifiable and the risk of privacy leakage still remains. 
Also, since the facial attributes are simply inherited from the target, the attributes from the source cannot be preserved to be suited for various needs.
To tackle this issue, most previous studies have taken either of the following two approaches: simply tweaking parts of the identity attributes or applying $k$-anonymity\footnote{The $k$-anonymity is a definition of the privacy security level.
Please refer to Newton~\etal~\shortcite{k_same} for details.}
to face images to ensure privacy at the cost of quality degradation.
Although these methods have paved a way for de-identification using deep learning, there is still room to improve. 
The risk of privacy leakage along with the challenge to control which facial attributes to preserve for data utility still remains for the first approach. Also, the blurry results from the averaged face of $k$ number of images reduce the image quality in both aesthetic and utility aspects for the second approach.

To tackle the issues, we have studied the under-explored challenges of securing a high level of privacy protection by achieving $k$-anonymity based on the $k$-same algorithm, while achieving the quality and controllability on the preservation of facial attributes.
In this work, we propose Facial Identity Controllable GAN (FICGAN), a highly secure and controllable face de-identification GAN model that learns to disentangle identity and non-identity attributes. 
Our method differentiates itself in two aspects: 
First, our method achieves $k$-anonymity for robust security by using the simple manifold $k$-same algorithm~\cite{k_same_sota}, which exploits mixing the $k$ number of different face images on the learned latent identity space as a target identity embedding. 
Second, the identity disentanglement and layer-wised generator enable detailed controllability on the degree of de-identification and also attribute preservation according to users' preference.
Numerous experiments demonstrate that our method can generate privacy-ensured de-identified face images with carefully controlled attributes from the source image to enhance data utility.
Furthermore, we provide ample discussions and analysis on the trade-off as well as the relationship between the degree of identity attributes and non-identity attributes of the human face.

\section{Related Work}
The traditional methods for face de-identification include black bar masking, blurring, and pixelization~\cite{deid_overview,deepfake_de_id}, which can aggressively destroy the facial information.
Thus, recent studies focus on preserving the facial attributes while effectively obfuscating the identity~\cite{live_face,acm}, by applying fusing~\cite{id_invert,fsgan,faceshifter,deepfake_de_id,deepfake1,deepfake2} and averaging~\cite{k_same,k_same_M,k_same_select,k_same_APFD,k_same_sota, averaging} the faces of the examples to remove the source identity and fill in with the target identity. 
Another method is to mask the entire facial area and synthesize a new face using inpainting~\cite{inpainting_deid,inpainting_deid2}.
The face swap based methods~\cite{fsgan,faceshifter,Swapitup,faceswap,faceswap2}
inevitably follow the attributes of the swapped face that may lead to another privacy leakage or malicious usage. 
Conditioning the landmarks~\cite{inpainting_deid,inpainting_deid2} of the face images for de-identification can be useful to maintain the facial expressions and head poses, but it is still limited to fully convey other physical traits; thus, those methods share the same limitation with the face swap based methods. 

To avoid identity leakage, it is more effective to utilize multiple images for de-identification. The $k$-same algorithms~\cite{k_same,k_same_select,k_same_M,k_same_APFD} for de-identification have been known to guarantee $k$-anonymity~\cite{k_same_sota}; however, the models employing $k$-same algorithms have not been able to generate high-quality outputs. 
However, our method achieves high-quality face de-identification with $k$-anonymity by using a manifold $k$-same algorithm, where $k$ indicates the number of different face images averaged on the learned latent space. 
Our model can generate highly realistic de-identified face images for enhanced data utility.
\begin{figure*}[t] 
    \vspace{-1.70em}
\centering
        \includegraphics[width=0.9\linewidth]{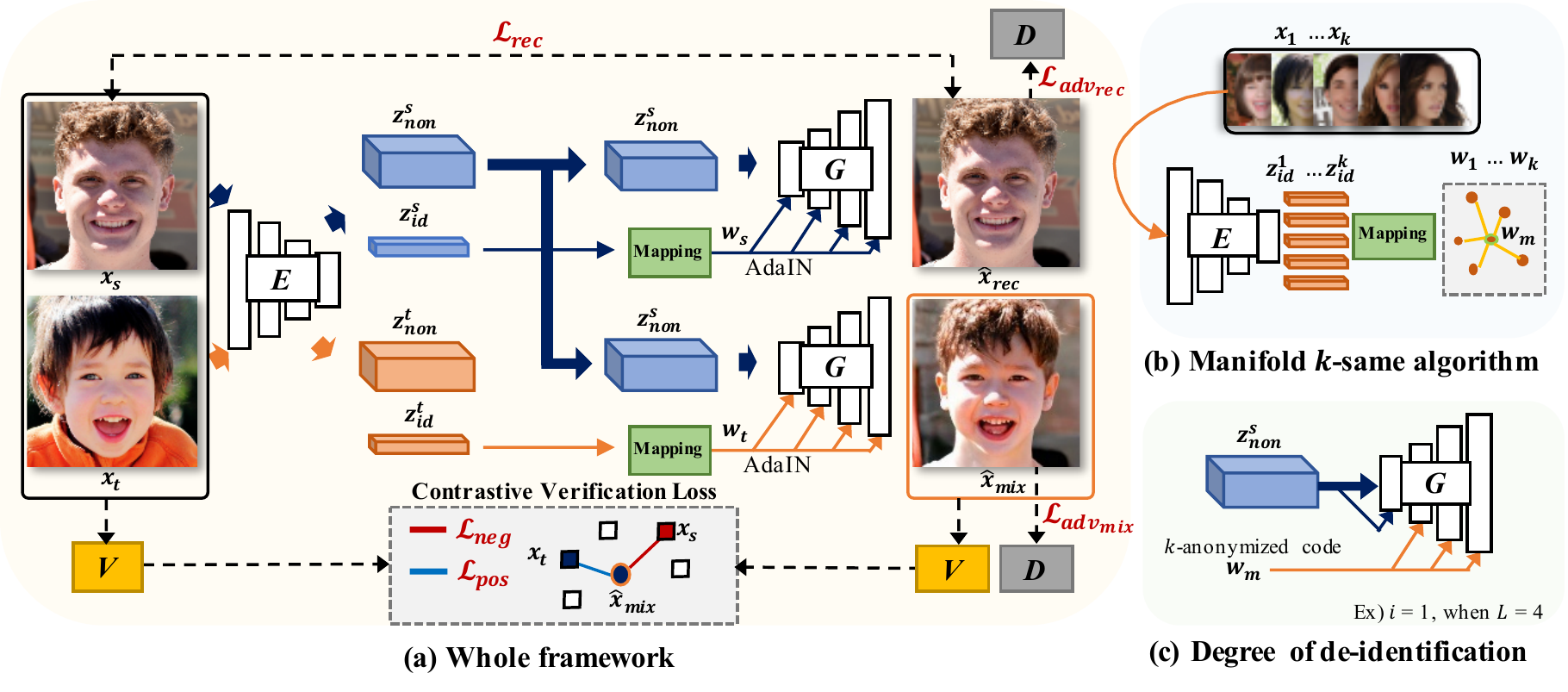}
\caption{\textbf{Overall framework.}
Our model learns two main tasks: (Top path) an auto-encoding based self-reconstruction task $\mathcal{L}_{Rec}$ with a sample $x_s$ and (Bottom path) an identity swapping task by contrastive learning, which facilitates the identity disentanglement (see (a)). 
We utilize the manifold $k$-same algorithm (see (b)) to control the degree of de-identification (see (c)) in the inference time.
}
    \label{fig:framework}
    \vspace{-1.70em}
\end{figure*}

The recent methods of advanced face de-identification have focused on disentangling the identity and non-identity attributes~\cite{acm, live_face, bao2018towards}. 
However, it is difficult to separate those attributes since they are rather mixed~\cite{face_recognition_critical,face_recognition_analisis,face_recognition_analisis_2,face_recognition_analisis_3}. This manifests as a trade-off between the degree of de-identification and preservation of attributes~\cite{xiao}, and any restriction on the de-identification tasks may yield undesirable face manipulation quality. Instead of restricting the region or attributes for de-identification, we design our model to control various properties for detailed control of the trade-off based on the user's preferences.

\section{Facial Identity Controllable GAN}
In this section, we provide detailed explanations on our face de-identification model, FICGAN.
We first describe the overall framework, then explain the loss functions used to train our model. Then we introduce four ways to utilize our model for various purposes.

\subsection{Overall Framework}
\label{sec:framework}
Our goal is to disentangle the identity (ID) and non-identity (non-ID) features to obtain detailed controllability on the trade-off between the degree of de-identification and preservation of facial attributes for enhanced data utility.
The encoder ($E$) is designed to disentangle the identity (ID) and non-identity (non-ID) attributes from the face images. 
For robust de-identification, our model mixes the ID attributes from the target image $x_t$ with the non-ID attributes from the source image $x_s$. Generator ($G$) de-identifies face images by mixing the features extracted from $x_s$ and $x_t$.
By incorporating the verification network ($V$) of SphereFace~\cite{sphereface} for loss function, our model can effectively disentangle the ID and non-ID attributes for robust face de-identification. 
The overall framework is shown in Figure~\ref{fig:framework}.

\paragraph{Encoder}
Embedding an image into the spatial and non-spatial codes with our careful loss design induces the disentanglement of the identity (ID) attributes from the non-identity (non-ID) attributes containing the spatial characteristics, such as the head pose, background, and local color scheme.
The encoder ($E$) takes the input images $x\sim\textbf{X}{\subset}\mathbb{R}^{H\times W\times 3}$ and extracts two types of features: the first is the ID latent code ($z_{id}$), which is a vector of size 512 for encoding the ID attributes; the second is the spatial latent code ($z_{non}$) titled the non-ID latent code, which is a 3-dimensional tensor of size $4\,{\times}\,4\,{\times}\,512$ encoding the non-identity attributes including the spatial information.
In contrast to the non-ID latent code containing the spatial information, such as the head pose, background, and local color scheme, the ID latent codes contain the non-spatial information, which make it easier to disentangle the ID and non-ID attributes.
It is similar to~\cite{swapping_autoencoder} that utilizes the latent codes from a different dimension to obtain the structural information.

\paragraph{Generator}
The ID and non-ID latent codes with the different dimensions from the encoder ($E$) are inserted as the input of the generator ($G$). 
The non-ID latent code is inserted into the generator in a feedforward manner to help generate the structural attributes, such as the head pose and expression, in the lower dimension of the image. 
In contrast, the ID latent code is processed by adaptive instance normalization (AdaIN)~\cite{adain1} to be adjusted to the generator in a layer-wise manner~\cite{stylegan}.
Before being inserted into the generator, the ID latent code $z_{id}$ is embedded to the intermediate latent vector $w$, which controls the ID of the generated faces.
It helps to locate the appropriate ID-space when applying the $k$-same algorithm in the domain of $z_{id}$ to generate high-quality de-identified images.

The mapping network consists of a series of multi-layer perceptrons (MLP). If the latent codes of ID ($z_{id}$) and non-ID ($z_{non}$) originate from a source image, $G$ generates a reconstructed image ($\hat{x}_{rec}$). In contrast, if $z_{id}$ and $z_{non}$ originate from different images, $G$ generates a de-identified image ($\hat{x}_{mix}$).

\paragraph{Face Verification Network}
We utilize the pre-trained verification network to effectively disentangle the ID attributes.
The face verification network ($V$) takes an image and extracts a feature vector of size 512. 
Because the verification networks extract discriminative features of facial identities, they are used for contrast learning to disentangle identities.

\subsection{FICGAN Losses}
\label{sec:losses}

\paragraph{Contrastive Verification Loss}
We disentangle it into two features $z_{id}$ and $z_{non}$, which allows independent modification of the facial attributes.
In order to disentangle the identity attributes from the input image $x$, contrastive learning~\cite{contrastive,contragan,consistency} is applied. 
From the input image $x$, the encoder $E$ separates the two latent codes $z_{id}$ and $z_{non}$. 
The two types of contrastive verification loss can be defined by the ID features of $x_s$, $x_t$, and mixed image $\hat{x}_{mix}$.
The embedding features are extracted by $V$, and the cosine similarity ($\langle\bullet,\bullet\rangle$) is used to calculate the similarity between the embedding features. 

The first type of loss to disassociate $\hat{x}_{mix}\,{=}\,G(z^t_{id},z^s_{non})$ from the $x_{s}$ in identity attributes can be defined as: 

\begin{equation}\label{eq:negative_loss}
        % \mathcal{L}_{neg} = \langle V(x_{s}), V(\hat{x}_{mix})\rangle.
        \mathcal{L}_{neg} = \langle V(x_{s}), V(\hat{x}_{mix})\rangle.
\end{equation}
where $z^t_{id}$ indicates the ID latent code disentangled from $x_t$, while $z^s_{non}$ indicates the spatial latent code of the non-identity attributes of $x_s$.
The second type of loss to associate $\hat{x}_{mix}$
close to $x_{t}$ in identity attributes can be defined as:

\begin{equation}\label{eq:positive_loss}
        % \mathcal{L}_{neg} = 1- \langle V(x_{t}), V(\hat{x}_{mix})\rangle.
        \mathcal{L}_{pos} = 1- \langle V(x_{t}), V(\hat{x}_{mix})\rangle.
\end{equation}

\paragraph{Reconstruction Loss}
The reconstruction loss is designed for the network to preserve all of the inserted information of the source image $x_s$.
Enabling to reconstruct $\hat{x}_{rec}$ from $x_s$, the reconstruction loss can be defined as below:

\begin{equation}\label{eq:rec_loss}
        \mathcal{L}_{rec}(E,G)=\mathbb{E}_{x_{s}\sim X}[||x-G(E(x))||_{2}].
\end{equation}
The reconstruction loss helps maintain the information of $x_s$ even after disentangling the ID and non-ID attributes.

\paragraph{Adversarial Loss}
The adversarial loss~\cite{goodfellow2014generative} is designed to ensure the visual quality of reconstructed images $\hat{x}_{rec}$ and mixed images $\hat{x}_{mix}$ as below:
\begin{eqnarray}\label{eq:adv_mix_loss,eq:adv_rec_loss}
\hspace{-5mm}&\mathcal{L}_{adv_{rec}}(E,G,D) =&\mathop{\mathbb{E}}\limits_{x_{s}\sim X}[-log(D(G(E(x_{s}))))],\\
\hspace{-5mm}&\mathcal{L}_{adv_{mix}}(E,G,D) =&\mathop{\mathbb{E}}\limits_{\substack{x_{s}, x_{t}\sim X,\\ x_{s}\neq x_{t}}}[-log(D(G(z^s_{non}, z^t_{id})))],\nonumber
\end{eqnarray}
It manages the degradation of image quality due to the blurriness caused by applying the $k$-same algorithm and the unnatural imagery due to a combination of $z^s_{non}$ and $z^t_{id}$.

\paragraph{Total Loss}
During an iteration of the training phase, the model performs both reconstruction and mixing.
The first is the reconstruction loss when $x_s=x_t$, and the second is the contrastive verification loss when $x_s\neq x_t$. 
Thus, the total loss can be defined as below:
\begin{equation}\label{eq:total_loss}
\begin{aligned}
        \mathcal{L}=\lambda_{1}\mathcal{L}_{rec}+\lambda_{2}(\mathcal{L}_{pos}+\mathcal{L}_{neg})+\lambda_{3}(\mathcal{L}_{adv_{rec}}+\mathcal{L}_{adv_{mix}}).\nonumber
\end{aligned}
\end{equation}

\subsection{Manipulation Methods at Inference Time}
\label{sec:usages}

\paragraph{Manifold $k$-Same Algorithm}
To obtain a high level of privacy protection, we apply $k$-same algorithm~\cite{k_same} in the latent space of our model, which is titled as \textit{manifold $k$-same algorithm}.
From the randomly selected $k$ facial images, the centroid of $\{w_1,w_2,\cdots,w_k\}$ is calculated by the encoder and the mapping network as shown in Figure~\ref{fig:framework}-(b). This centroid vector (or $k$-anonymized code; $w_m$) satisfies the $k$-same algorithm because the properties of $k$ are combined.
Based on $w_m$, our single model can perform de-identification in four other ways with securing $k$-anonymity, depending on the purpose of use as follows.

\paragraph{Controlling Degree of De-identification} 
We leverage the layer-wise controllability offered by StyleGAN~\cite{stylegan} architecture. 
Unlike StyleGAN that simply mixes up the styles in images, we mix up the ID attributes of the source and target images in a layer-wise manner for a fine-grained control to manage the trade-off between the degree of de-identification and preservation of attributes for enhanced data utility. 
In this way, we can observe the transition resulting from adjusting the identity-related characteristics in facial attributes.
First, we can control the degree of de-identification by selecting the index of layer ($i$), which is the point of standards to adjust the amount of $w_m$ and $w_s$ obtained from the source image $x_s$. 
To be specific, we insert $w_s$ from the first to $i$-th layers and $w_m$ from $i+1$-th to $L$-th layers. When $i=0$, mean identity $w_m$ is inserted to all layers, thus the source image is strongly de-identified (strong de-ID). In contrast, when $i=L-1$, $w_m$ is applied only to the highest layer, thus leading to a weak de-identification (weak de-ID).
An example (i=1, when L=4) of controlling the degree of de-identification is shown in Figure~\ref{fig:framework}-(c).
The greater the $i$ is, the less the degree of de-identification is, which leads to more attributes preserved to enable a fine-grained control. 
We provide detailed explanations on the trade-off between the identity attributes and non-identity attributes according to the adjustment of $i$ in the paragraph of `Trade-off between De-identification and Attribute Preservation' under Experiments.

\begin{figure*}[]
\centering
        \includegraphics[width=0.9\linewidth]{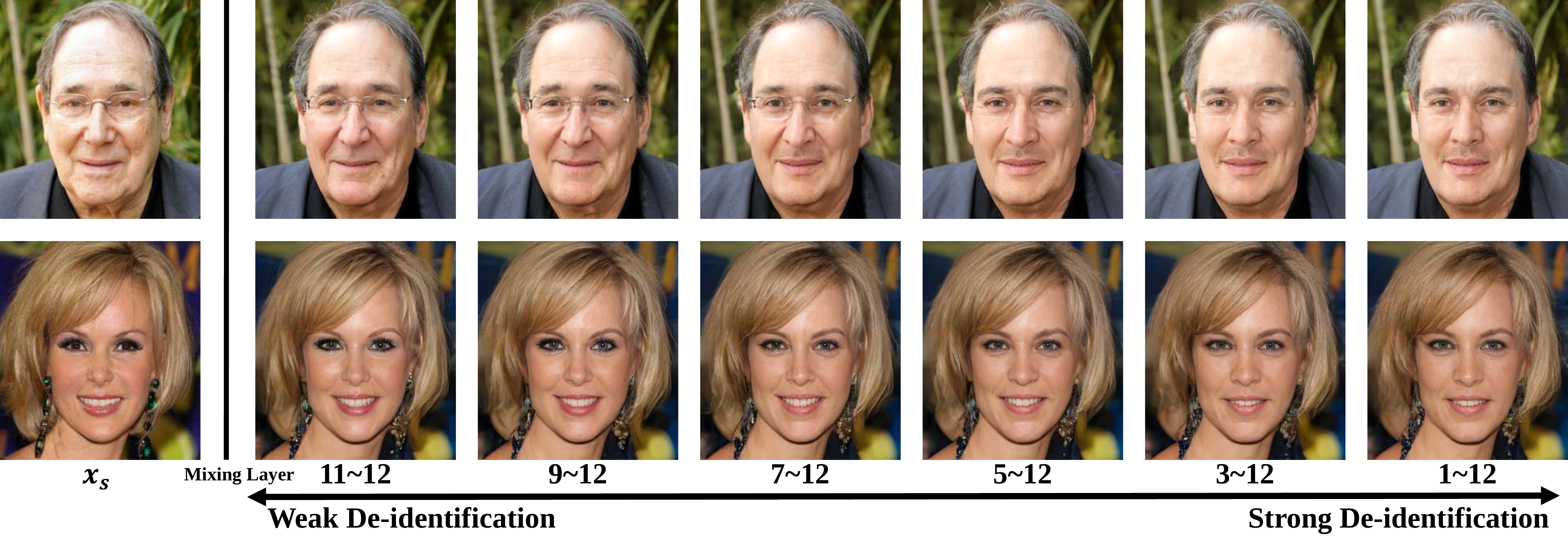}
\caption{\textbf{Layer-wise de-identification performance.} The results show progressions from weaker to stronger de-identification. We control the degree of de-identification by selecting the layers to feed the $k$-anonymized vector $w_m$. It shows that feeding the code $w_m$ from the highest layer to the lower layers induces stronger de-identification, and vice versa.}
    \vspace{-0.7mm}
\label{fig:layer}
    \vspace{-2.7mm}
\end{figure*}

\begin{figure}[t]
\centering
        \includegraphics[width=1.0\linewidth]{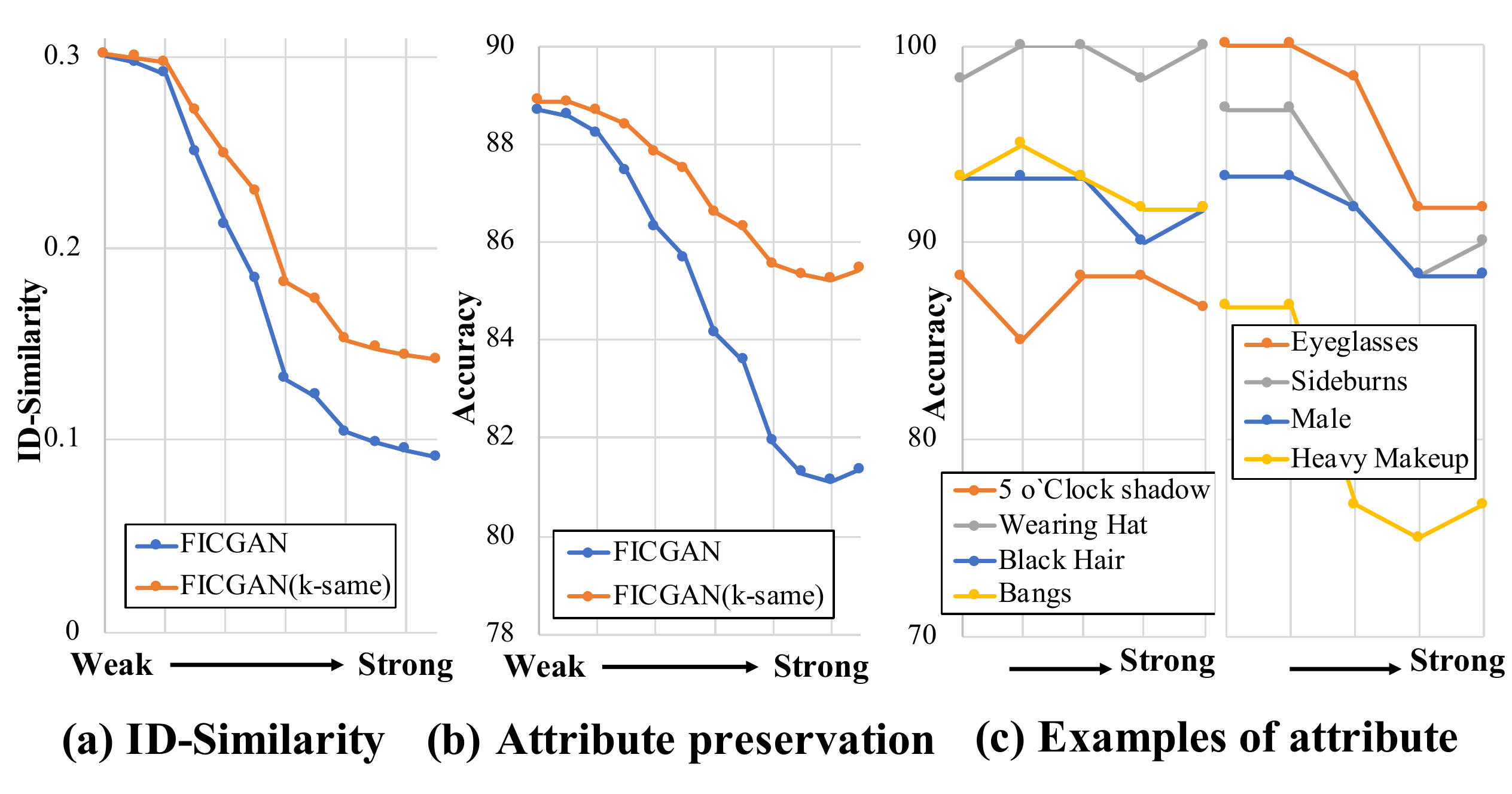}
\caption{\textbf{Quantitative results of FICGAN.}
(a) ID-Similarity and (b) preservation accuracy of total attributes 
according to the degree of de-identification. (c) shows preservation accuracy for example of attributes.}
    \label{fig:quantitative_de_id}
    \vspace{-0.5em}
\end{figure}

\paragraph{Attribute Control}
We can control specific attributes of the image according to the user's preference.
Loss of specific attributes is inevitable due to a trade-off between the identity and non-identity attributes~\cite{k_same_select}. However, we opt to control the loss by employing $w'_{m}$, the mean vector of specific attributes obtained from a group of images with the same attributes. Using $w'_{m}$, we can either preserve or insert specific attributes for fine-grained control of de-identification.
We provide detailed explanations on how our model controls the preservation of attributes as desired in the paragraph of `Controllability on Facial Attributes' under Experiments.

\paragraph{Example-based Control}
Our model can control the direction of de-identification using the examplar of face images, therefore achieving data diversity.
The increase of $k$ in $k$-anonymity has a twofold effect: it can guarantee a greater degree of anonymity for privacy protection, but can also lead to excessive generalization of mean facial identity, resulting in identical faces.
However, data diversity can be required in some cases. For example, if multiple people are presented in the same scene, their de-identified faces should be different from each other. Also, it is important to consider diversity in race, ethnicity, gender, and age to fully represent a broad range of faces.

For such cases, we can generate diverse images of de-identified faces by controlling the direction and the degree of de-identification through the interpolation between $w_t$ and  $w_m$. The $k$-anonymized code for direction, $w_d$, can be calculated by employing the scaling factor $\alpha$ as below:
\begin{equation}\label{eq:direction_w}
        w_d = w_m + \alpha\cdot (w_t {-} w_m).
\end{equation}
It utilizes the linear interpolation $w'$ between $w_m$ and the latent space of a specific target image $x_t$. 
Diverse images of de-identification are provided in the paragraph of 'Diversity of De-identified Images' under Experiments. 

\paragraph{Identity Swapping}
Lastly, target ID code $w_t$ is employed instead of $w_m$ for identity swapping to switch to the target identity.
The results are illustrated in Fig.~\ref{fig:id_swapping}.

\section{Experiments}
\begin{figure*}[t]
\centering
        \includegraphics[width=0.88\linewidth]{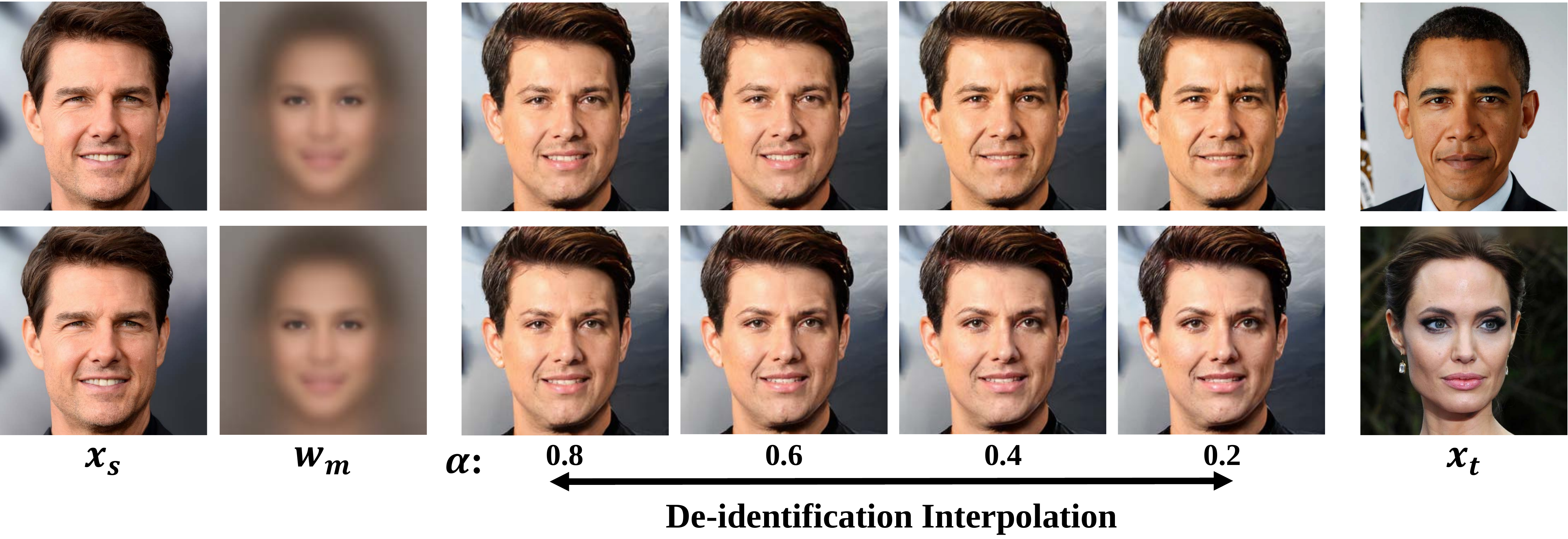}
\caption{\textbf{The results of our example-based control with $k$-same algorithm.} Our method can manipulate the degree of de-identification as well as the direction to be manipulated with a target identity $w_t$ from a $k$-anonymized code $w_m$, \ie, $w_t {-} w_m$. The degree is controlled by a scaling factor $\alpha$ by changing the identity feature as $w_m + \alpha\cdot (w_t {-} w_m)$.
}
    \label{fig:diverse}
    \vspace{-0.6cm}
    % \vspace{-1.5em}
\end{figure*}

\begin{figure}[t]
\centering
        \includegraphics[width=1.0\linewidth]{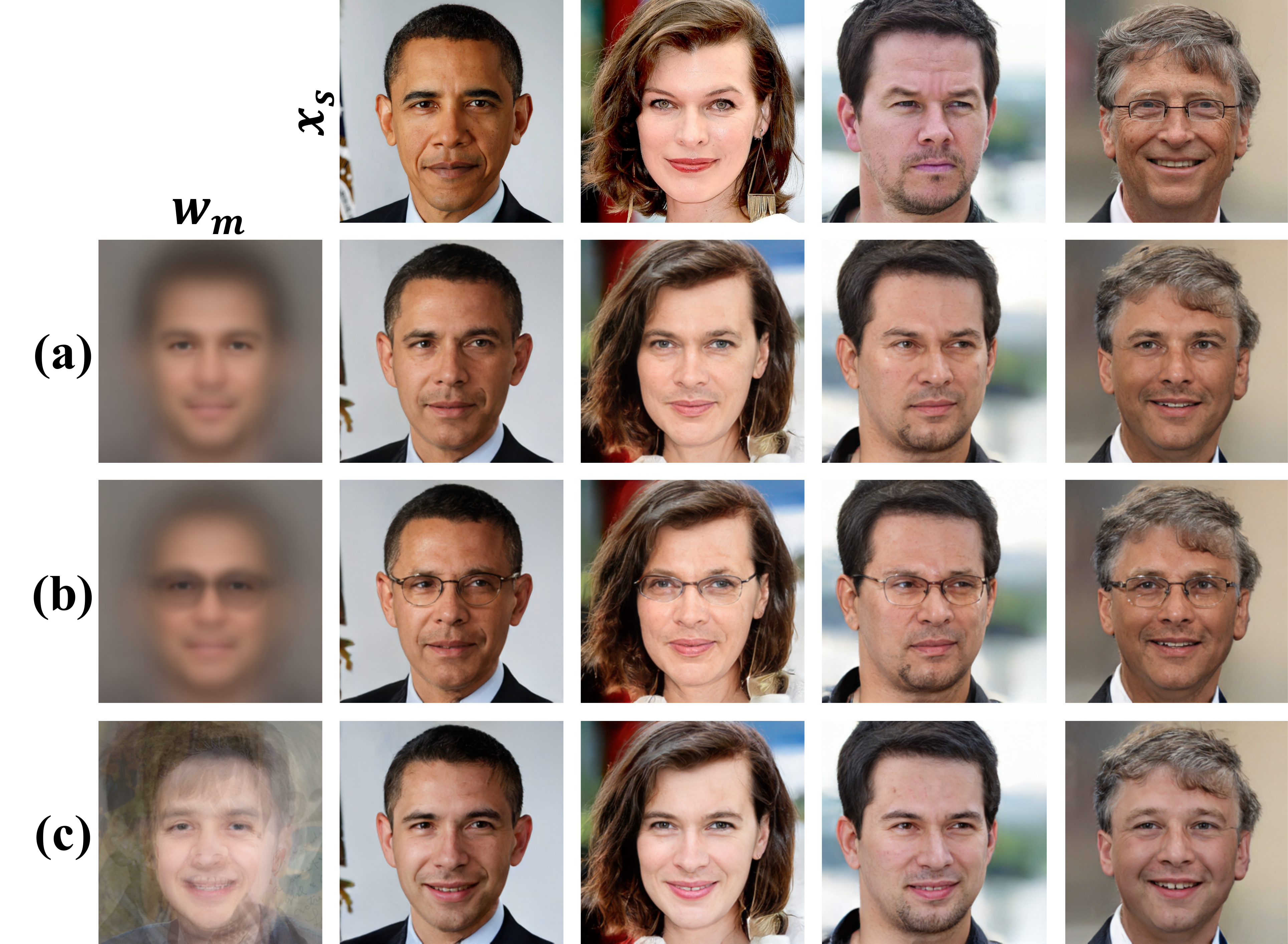}
\caption{\textbf{Our attribute control results with $k$-same algorithm.} Our method can de-identify with controlling specific attributes, \eg, (a) male, (b) male wearing glasses, and (c) smiling people with pale skin.
Specific attribute effects can be attained by the $w_m'$ encoding target attributes to control. The code $w_m'$ is obtained by averaging ID latent codes of the $k$ number of subjects sharing the target attributes.
}
    \label{fig:control}
    \vspace{-0.6cm}
\end{figure}

\begin{figure}[t]
\centering
        \includegraphics[width=0.92\linewidth]{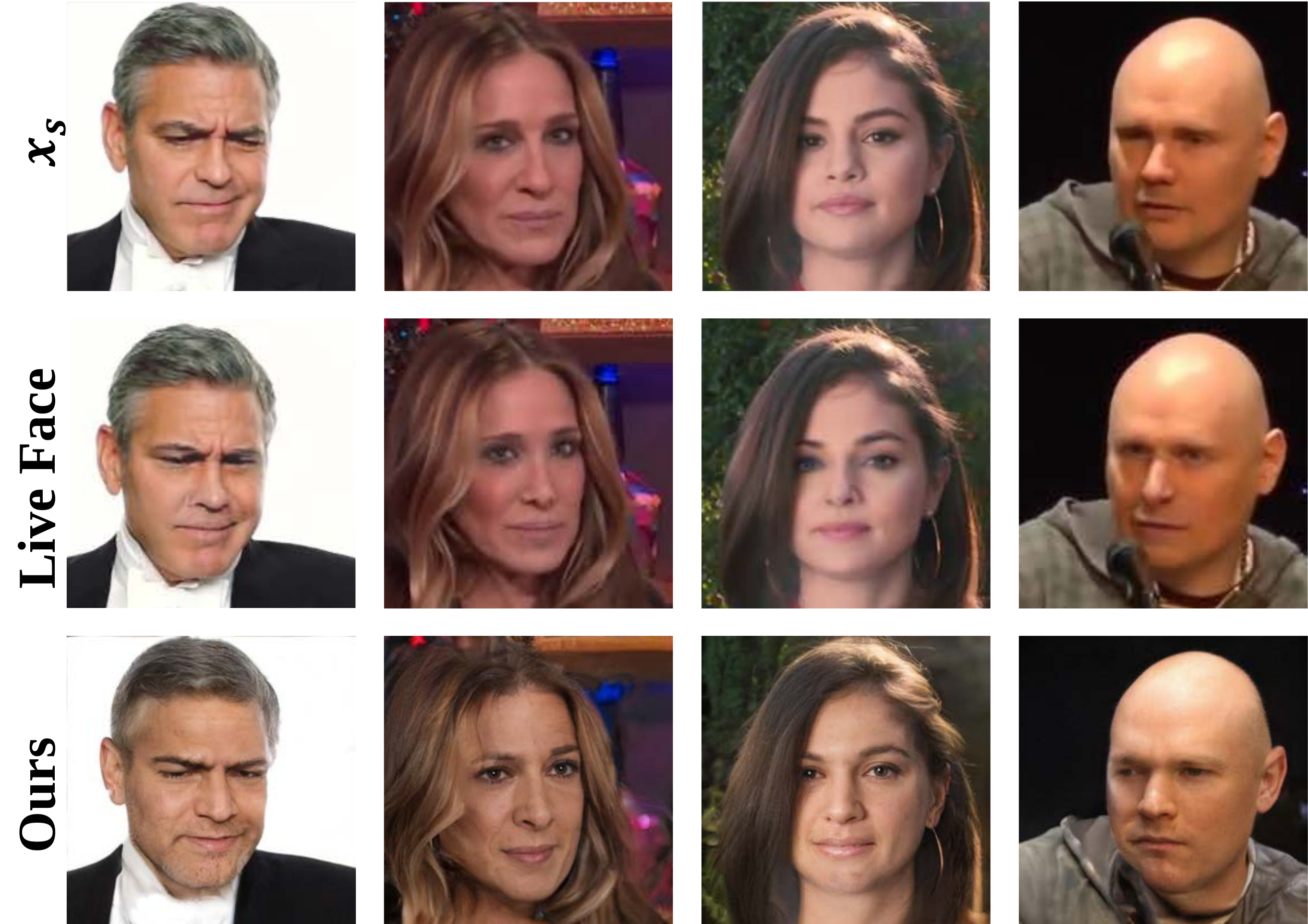}
\caption{\textbf{Comparison to Live Face} Compared to our method, Live Face shows marginal de-identification effects that might be ineffective for human subjects.
}
    \label{fig:de_id}
    \vspace{-0.5em}
    % \vspace{-1.5em}
\end{figure}

\subsection{Dataset}
For training, we use 60,000 images out of 70,000 images in FFHQ~\cite{stylegan}. 
For testing, we use the remaining 10,000 images of FFHQ and the unseen CelebA-HQ dataset. To compare with the existing methods, we use celebrity face images from ALAE~\cite{alae}, IDInvert~\cite{id_invert}, and Liveface~\cite{live_face}.

\subsection{Evaluation Metrics}

For quantitative analysis, we use three types of metrics. First, we measure the Fr\'echet Inception Distance (FID)~\cite{fid} score to evaluate the quality of de-identified images.
Then, to test the effectiveness of de-identification, we measure the cosine similarity between the identity features of the source image $x_s$ and the de-identified image $\hat{x}$. 
For a fair comparison of the identity similarity, we use the pre-trained ArcFace~\cite{arcface}, the face verification network excluded from our training.
To evaluate whether the attributes are well preserved, we train a multi-class classifier with CelebA-HQ~\cite{celebahq} which contains images labeled with 40 attributes. We use 27,000 images from CelebA-HQ for training and the remaining 3,000 images for testing.
The average performance of the classifier for 40 attributes is 90\% on the test set.

\subsection{Trade-off between De-identification and Attribute Preservation}
~\label{sec:trade_off}
In this paragraph, we measure the degree of de-identification and the preservation of the attributes according to the layer where $w_m$ is applied. Consecutively inserting $w_m$ from the lowest to highest layers leads to greater application of $k$-anonymized identity, resulting in strong de-ID of the source face ($x_s$). In contrast, inserting $w_m$ only to the highest layers leads to greater preservation of the facial features in $x_s$, resulting in weak de-ID as shown in Figure~\ref{fig:layer}.
Figure~\ref{fig:quantitative_de_id}-(a,b) shows the performance of ID-similarity and attribute preservation when the $k$-same algorithm ($w_m$) or an arbitrary ID~($w_t$) is applied. Also, Figure~\ref{fig:quantitative_de_id}-(c) shows the accuracy of the attribute to evaluate the preservation of each attribute. 
This quantitatively explains the relationship between the degree of de-identification and preservation of the attributes for each layer.

As shown in Figure~\ref{fig:quantitative_de_id}-(a), when strong de-ID is applied ID-similarity decreases and it is favorable in terms of de-identification, but the accuracy of the attributes also decreases, which may not be favorable depending on user's preference. 
As shown in Figure~\ref{fig:layer}, the glasses are removed (first row) while bangs (second row) are preserved when strong de-ID is applied. It can be also observed in Figure~\ref{fig:quantitative_de_id}-(c).
This is because glasses play an important role in recognizing the identity of an individual. As illustrated, the stronger the degree of de-identification is, the weaker the attributes are preserved.
A deep analysis of the ID and non-ID attributes per layer of the network is provided in supplementary materials.

\subsection{Controllability on Facial Attributes}~\label{sec:control}
Figure~\ref{fig:control} illustrates how the facial attributes can be controlled using the averaged attribute code $w_m$. 
In the images of the last column in Figure~\ref{fig:control}, applying the attributes of (a) male to the source image $x_s$ effectively preserves the male attributes but removes the glasses of $x_s$, due to their high correlation to identity. This can be improved by applying the attributes of (b) male wearing glasses. In addition, attributes such as male, smiling, and pale-skinned can be added to the source image to generate controlled de-identified images.

Our model not only controls the attributes of face images using the averaged attributes with high complexity, but also performs de-identification suitable for k-anonymity. 
More diverse results of the k-same algorithm applied to other models are provided in supplementary materials.% Appendix~\ref{sec:k-same}. 

\subsection{Diversity of De-identified Images}~\label{sec:diversity} 
Figure~\ref{fig:diverse} shows that example-based control with equation~\ref{eq:direction_w} can manipulate the degree of de-identification as well as the direction to be manipulated with a target identity $w_t$ from a $k$-anonymized code $w_m$. 
Although the same mean attribute $w_m$ is used in Figure~\ref{fig:diverse}, the de-identified results gradually become to look more like the corresponding $w_t$ based on the interpolation, as shown in the eyebrows and skin color. 
Moreover, it guarantees anonymity even for the interpolation between $w_t$ and $w_m$, since it uses an arbitrary $z^t_{id}$ other than those that have been already disentangled from $z^s_{non}$. 

% Please add the following required packages to your document preamble:
% \usepackage{multirow}
\begin{table}[]
\centering
\resizebox{1.00\linewidth}{!}{%
  \begin{tabular}{@{\,}c@{\quad}c@{\quad}c@{\quad}c@{\,}}
  \hline
  Model  &$\downarrow$ FID & $\downarrow$ ID Similarity & $\uparrow$ Attribute \\ \hline
  FSGAN~\cite{fsgan} &  45.62 & 0.21 &79.92\\
  IDDis~\cite{acm} &  46.48 & 0.01 & 74.31\\
  FICGAN & 20.45 & 0.09 & 81.35 \\
  \hline
  FSGAN~\cite{fsgan} + $k$-same & 63.24 & 0.33 & 83.56\\
  IDDis~\cite{acm} + $k$-same & 131.23 & 0.01 & 78.06\\
  FICGAN + Manifold $k$-same (Ours) & 22.25 & 0.14 & 85.44\\ 
  \hline
  \end{tabular}
}
\caption{\textbf{Quantitative comparison with other models. } 
}

\label{tab:de_id_result}
\vspace{-0.5cm}
\end{table}

\subsection{Comparison to Previous Literature}~\label{sec:comparison} %Comparison with Previous Works.

\begin{figure}[t]
\centering
        \includegraphics[width=0.92\linewidth]{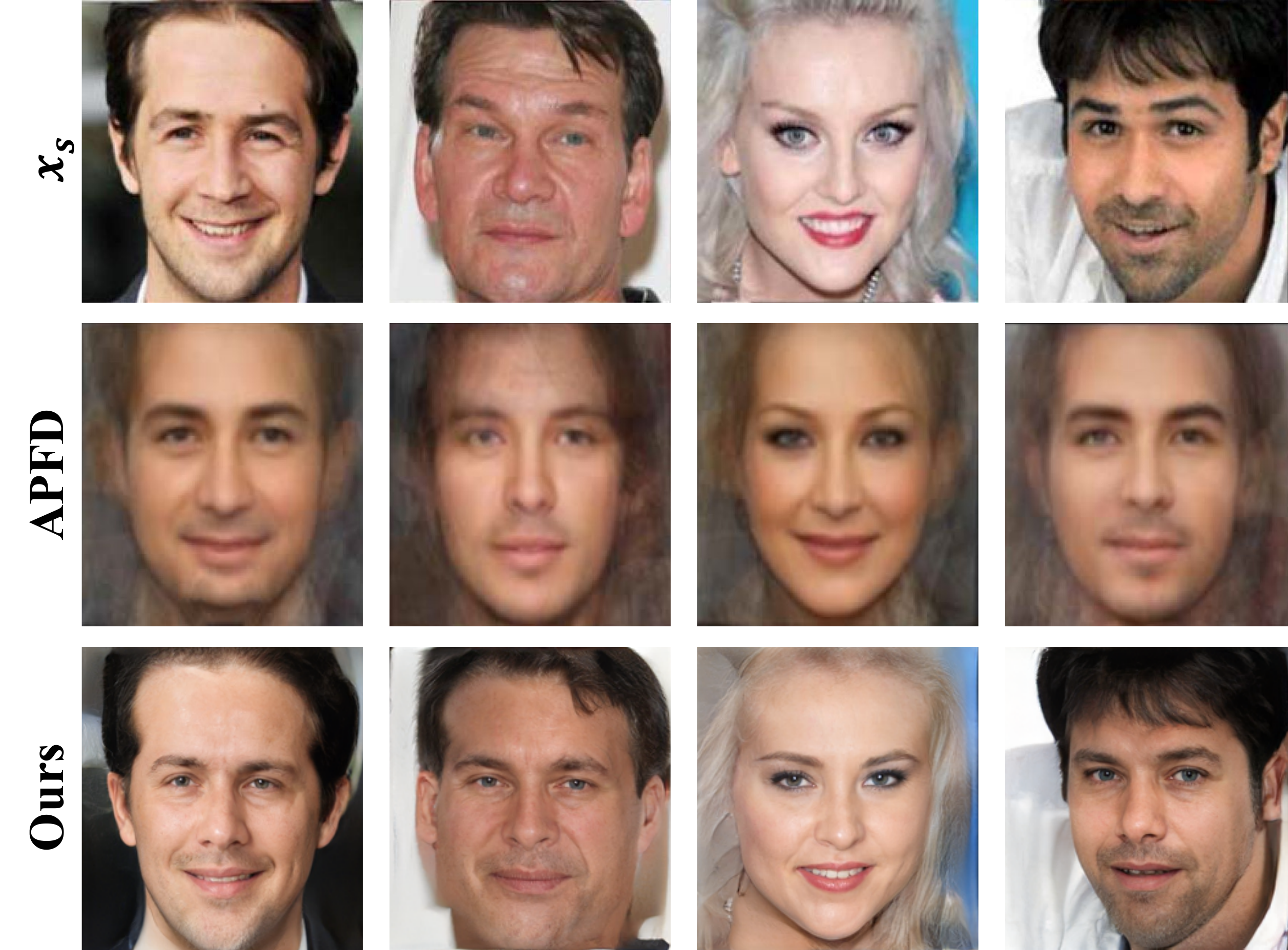}
\caption{\textbf{Comparison to the APFD.} 
APFD shows weak de-identification performance and low-quality with destroying many attributes in the sources $x_s$, while our method de-identifies seamlessly with preserving most of the non-identity attributes.
        }
    \label{fig:k_same_result}
    \vspace{-0.5em}
    % \vspace{-1.5em}
\end{figure}
\begin{figure}[t]
\centering
        \includegraphics[width=0.92\linewidth]{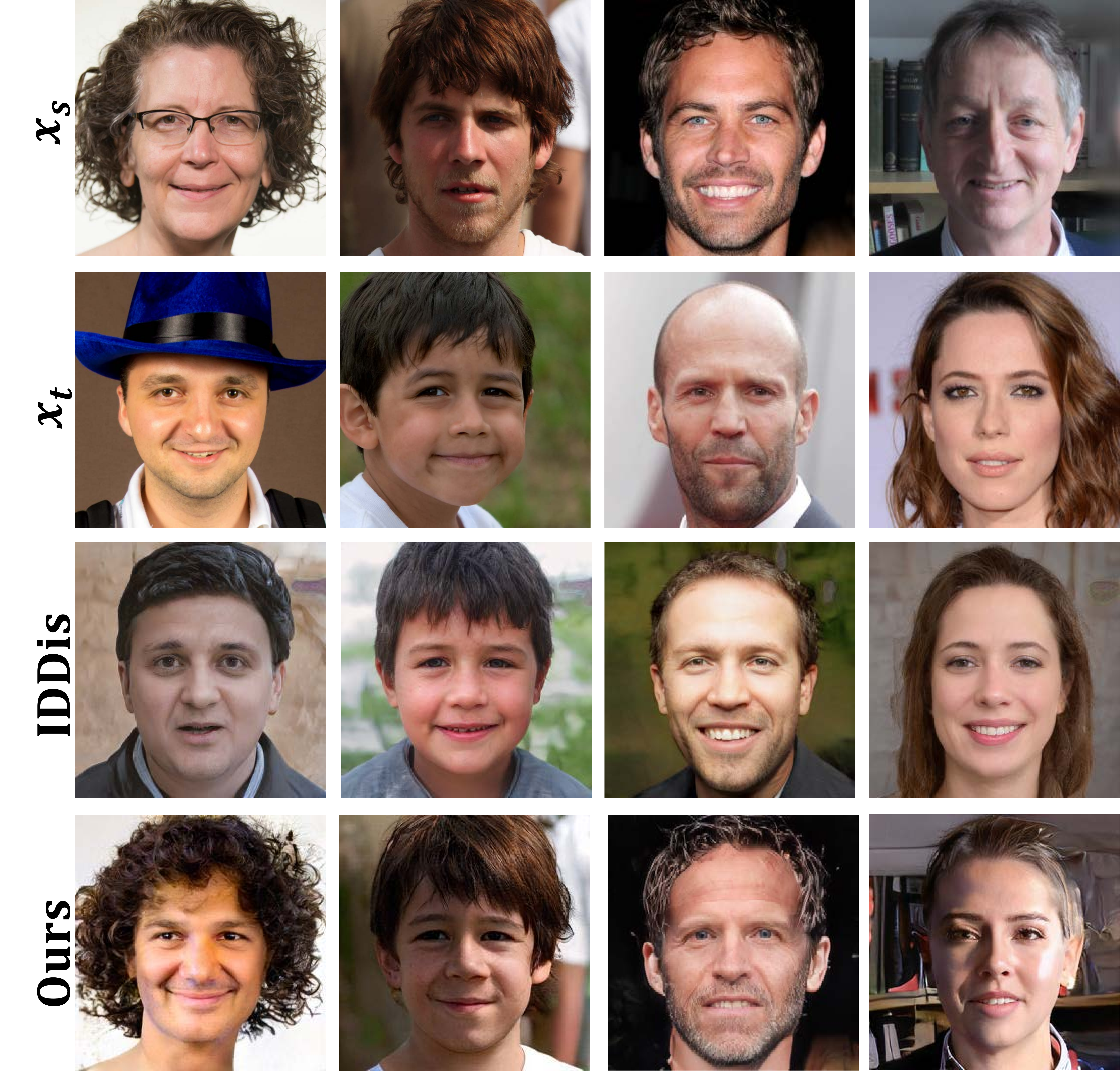}
\caption{\textbf{Comparison to Identity Swapping models.}The results are obtained by swapping the identity of the target $x_t$ on the source image $x_s$ with IDDis and ours.
The first two left and last two right examples of $x_s$ and $x_t$ are from the unseen FFHQ and IDInvert, respectively.
}
    \label{fig:id_swapping} 
    \vspace{-1.0em}
\end{figure}

In this paragraph, we conduct quantitative and qualitative analyses on our model compared to the previous literature. Figure~\ref{fig:de_id} illustrates the de-identified images based on the images used in Live Face~\cite{live_face}. The results of our model preserve the head pose, expression, skin tone, and hairstyle of the face of $x_s$, while successfully obfuscating the identity. Also, Figure~\ref{fig:k_same_result} illustrates the results of our model compared to APFD~\cite{k_same_sota}, which applied k-same algorithm. The results of APFD are blurry and unnatural with the disappeared head pose, while our model successfully performs de-identification even with the k-same algorithm applied.
Figure~\ref{fig:id_swapping} illustrates the de-identified images of our model compared to those of other face-swapping models. ID-Dis~\cite{acm} only preserves the head pose of the source image and disregards the rest of the non-identity attributes. 

By effectively managing the trade-off between the ID and non-ID attributes, our model achieves exceptional performance in preserving the non-ID attributes up to a level where they are almost identical when evaluated with the CelebA-HQ test set. The results in Table~\ref{tab:de_id_result} show that our method produces the most realistic face images, as shown by the FID score.
For the de-identification performance  measured by ID similarity, our method is in the second place after IDDis. However, as shown in the attribute measure in this table and will be shown in Figure~\ref{fig:id_swapping}, our method preserves the high-quality source attributes, while IDDis does not. Besides, the high FID fluctuations at ``FSGAN+$k$-same'' and ``IDDis+$k$-same'' imply that it is difficult to extend those methods to $k$-anonymity.
More experiments on the comparative analysis are provided in supplementary materials.
\vspace{-0.5em}\subsection{Ablation Study}~\label{sec:ablation}\vspace{-1.0em}

\begin{figure}[t]
\centering
        \includegraphics[width=0.99\linewidth]{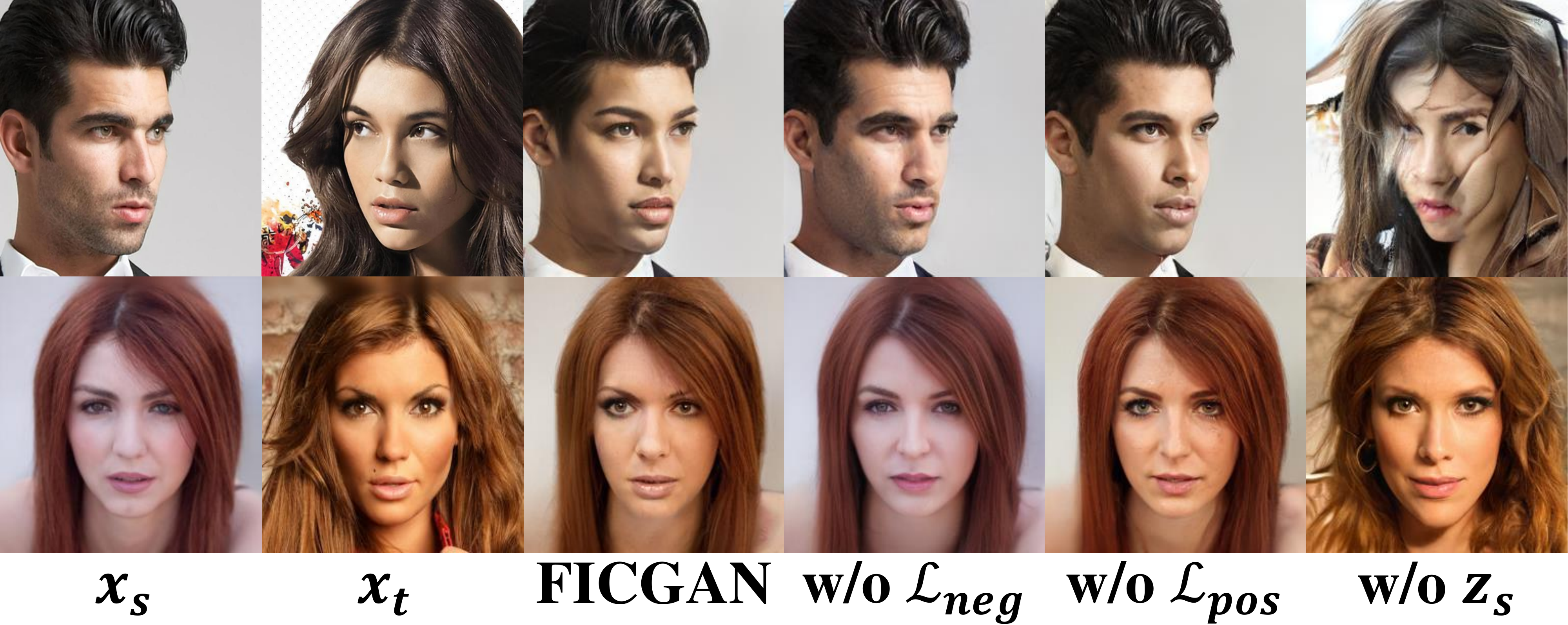}
\caption{\textbf{Qualitative analysis of $\mathcal{L}_{neg}$, $\mathcal{L}_{pos}$, and $z_s$ using Identity Swapping.} }
    \label{fig:ablation} 
    \vspace{-1em}
\end{figure}

\begin{table}[t]
\centering
\scriptsize
  \begin{tabular}{ccc}
  \hline
  Model & $\downarrow$ Similarity & $\uparrow$ Attribution \\ \hline
  FICGAN  & 0.091 & 81.35 \\
  w/o Spatial code ($z_s$) & 0.008 & 74.49\\
  w/o Positive loss ($L_{pos}$) & 0.192 & 85.76 \\
  w/o Negative loss ($L_{neg}$) & 0.312 &  89.02 \\ \hline
  FICGAN  & 0.142 & 85.44 \\ 
  w/o Spatial code ($z_s$) + k-same & 0.001 & 78.09\\
  w/o Positive loss ($L_{pos}$) + k-same & 0.189 & 87.01 \\
  w/o Negative loss ($L_{neg}$) + k-same & 0.313 &  89.05 \\
  \hline
  \end{tabular}
\caption{\textbf{Ablation Study.} } 
\label{tab:ablation_result}
\vspace{-0.5em}
\end{table}

Figure~\ref{fig:ablation} shows the effect of the proposed spatial code ($z_s$) and two types of contrastive verification loss ($\mathcal{L}_{neg}$, $\mathcal{L}_{pos}$). If $\mathcal{L}_{neg}$ is absent, the model tends to maintain the identity of the source image, which indicates that $\mathcal{L}_{neg}$ improves the quality of attribute disentanglement. Though less noticeable, $\mathcal{L}_{pos}$ makes the swapped identity appear closer to $x_t$. If $z_s$ is replaced with the constant vector, it can result in the overall facial features distorted and the non-identity attributes of $x_s$ vanished. Table~\ref{tab:ablation_result} also shows more cases of ablation study. Without $z_s$, it results in the lowest similarity but the least amount of remaining attributes, which indicates the importance of $z_s$ in preserving the attributes in $x_s$. Also, it shows that employing both of $\mathcal{L}_{neg}$ and $\mathcal{L}_{pos}$ results in a higher quality of de-identification.

\section{Conclusion}
We propose FICGAN that generates high-quality de-identified face images with robust privacy protection satisfying $k$-anonymity. 
By enabling detailed controllability on the trade-off between the degree of de-identification and preservation of facial attributes, our method can generate diverse de-identified images without sacrificing image quality. 
We empirically demonstrate that our method outperforms other methods in various scenarios. Based on the analysis and discussions on the relationship between the ID and non-ID attributes, our method opens up a new door for future work and even new promising applications, \eg, a media director can protect the identity of interviewees while controlling the facial attributes in detail for various needs.

% \newpage
{\small
\bibliography{aaai}
}
\clearpage
\newpage
\setcounter{table}{0}
\renewcommand{\thetable}{\Alph{table}}
\setcounter{figure}{0}
\renewcommand{\thefigure}{\Alph{figure}}
\setcounter{section}{0}
\renewcommand\thesection{S\Alph{section}}

\twocolumn[
\centering
\Large
\textbf{FICGAN: Facial Identity Controllable GAN for De-identification} \\
\vspace{0.5em}-- Supplementary Material -- \\
\vspace{1.0em}

\begin{center}
\centering
{\includegraphics[width=0.99\linewidth]{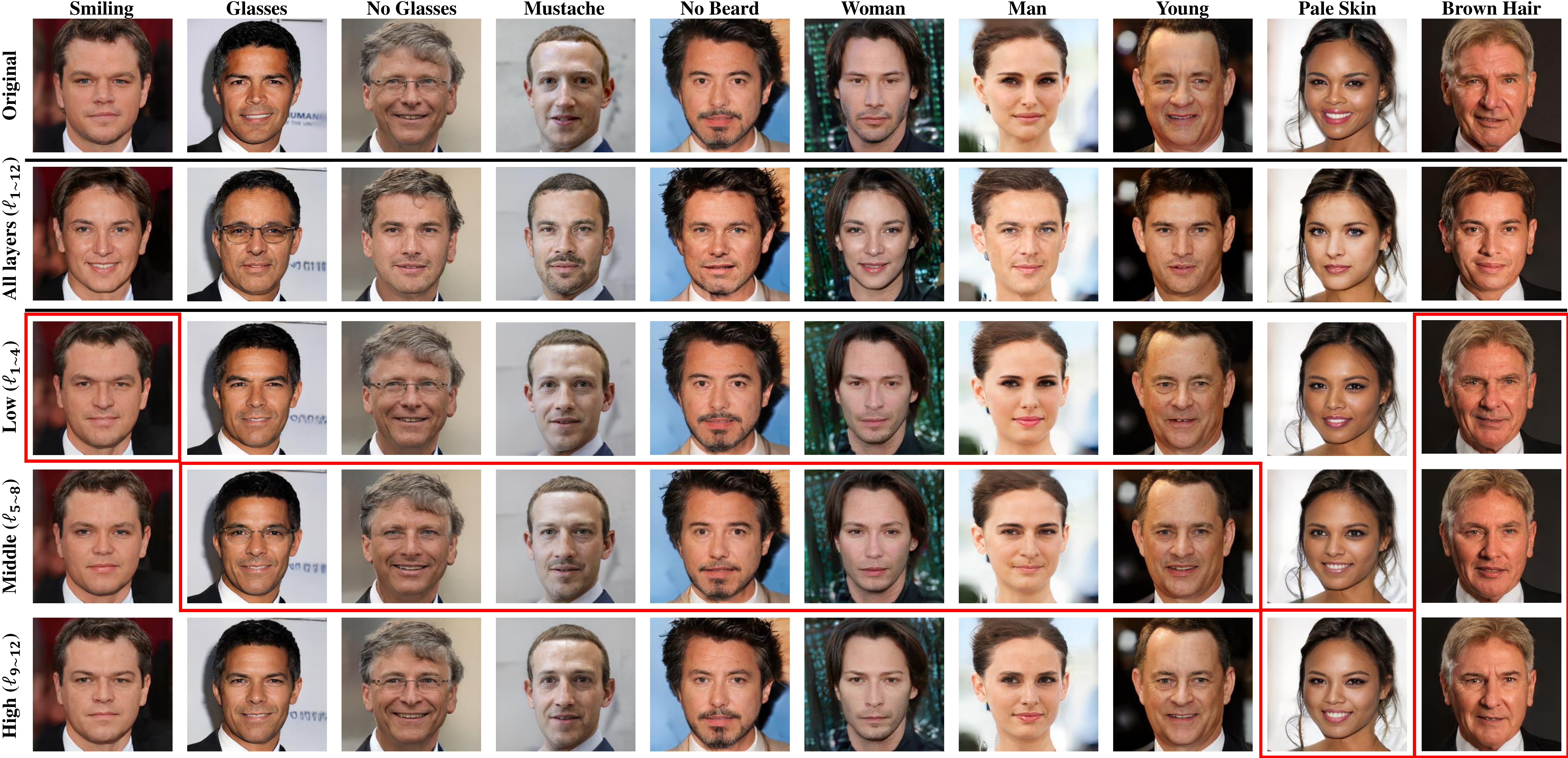}}
\captionof{figure}{\textbf{Layer-wise control of attributes for De-identification.} 
The de-identified results of layer-wise control of attributes, such as smiling, glasses, mustache, gender, age, pale skin, and hair.Starting from the original images, we have inserted $w_m$ into all layers ($l_{1\sim12}$: layers 1 to 12), and low layers ($l_{1\sim4}$: layers 1 to 4), middle layers ($l_{5\sim8}$: layers 5 to 8), and high layers ($l_{9\sim12}$: layers 9 to 12), respectively. The red boxes indicate the de-identified images in which each attribute is most reflected. In case of brown hair, the color changes from blonde to brown only if $w_m$ is inserted into all layers. 
}
    \label{fig:layer_w}
    % \vspace{-1.5em}
\end{center}
]
\appendix

\begin{figure*}[t]
\centering
        \includegraphics[width=0.8\linewidth]{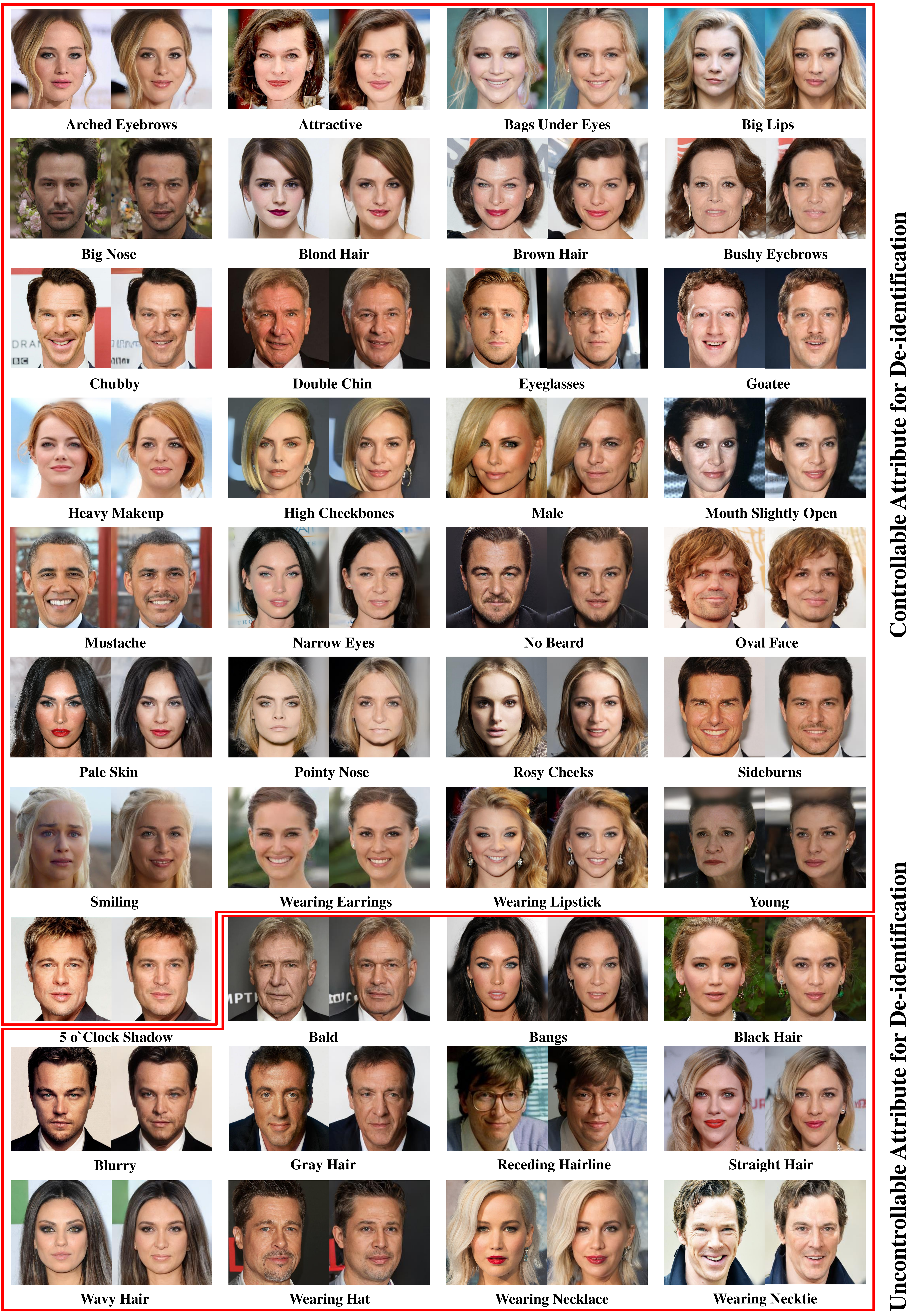}
\caption{\textbf{The controllable and uncontrollable attributes for de-identification.} The figure illustrates the original image and the modified image for 40 attributes of CelebA-HQ~\cite{celebahq}, respectively. It is divided by the red boxes indicating the controllable and uncontrollable attributes for de-identification, according to the relation to identity considered by the model.
}
    \label{fig:controllable_att}
    % \vspace{-1.5em}
\end{figure*}

\begin{figure}[t]
\centering
        \includegraphics[width=0.99\linewidth]{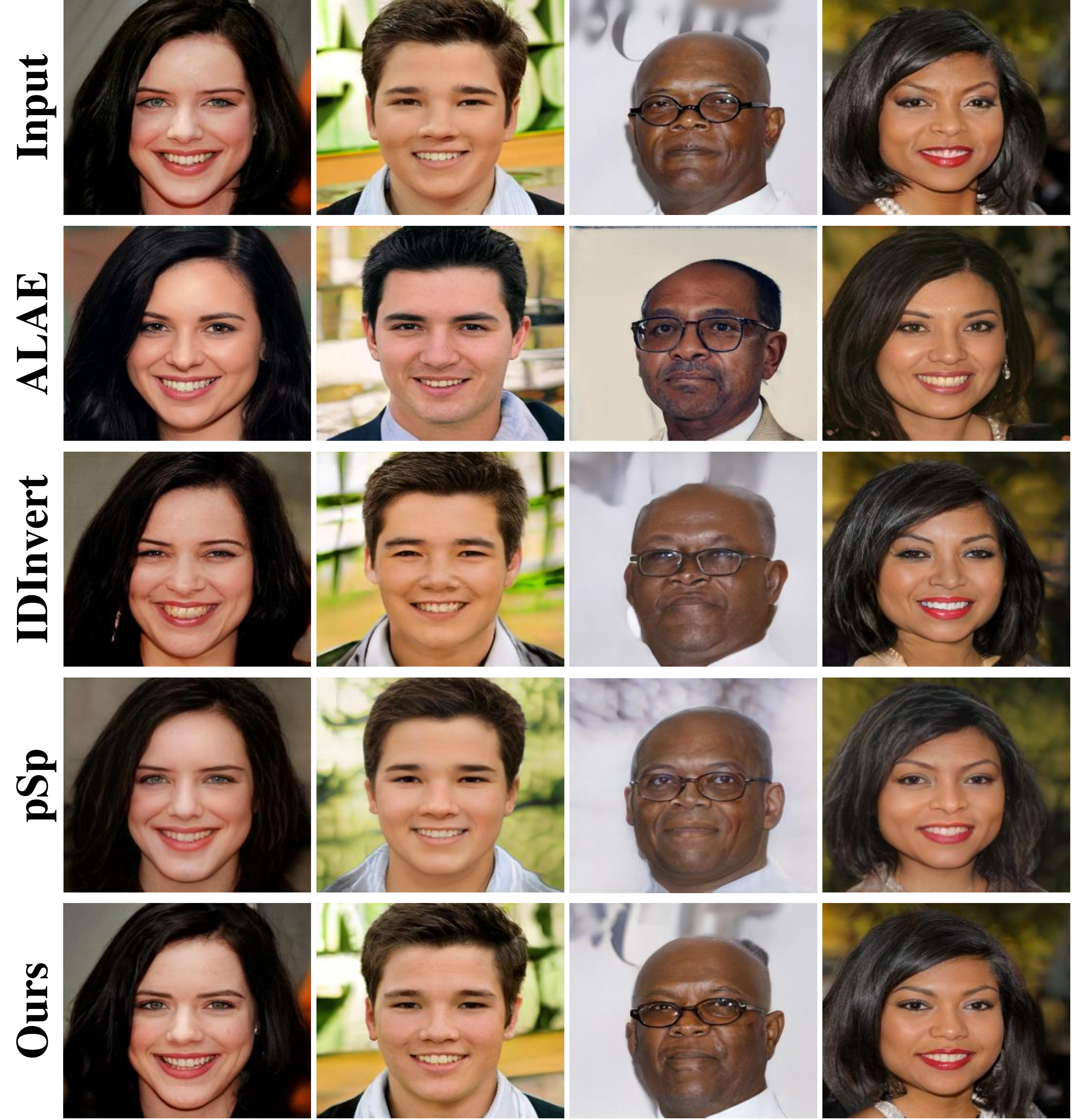}
\caption{\textbf{CelebA-HQ reconstruction.} CelebA-HQ reconstruction of unseen samples. Top row : original images. Second row: ALAE~\cite{alae}, Third row: IDInvert~\cite{id_invert}, Fourth row: pSp~\cite{psp}, and Last row: ours.}
    \label{fig:recon}
\end{figure}

\begin{figure}[t]
\centering
    \includegraphics[width=0.99\linewidth]{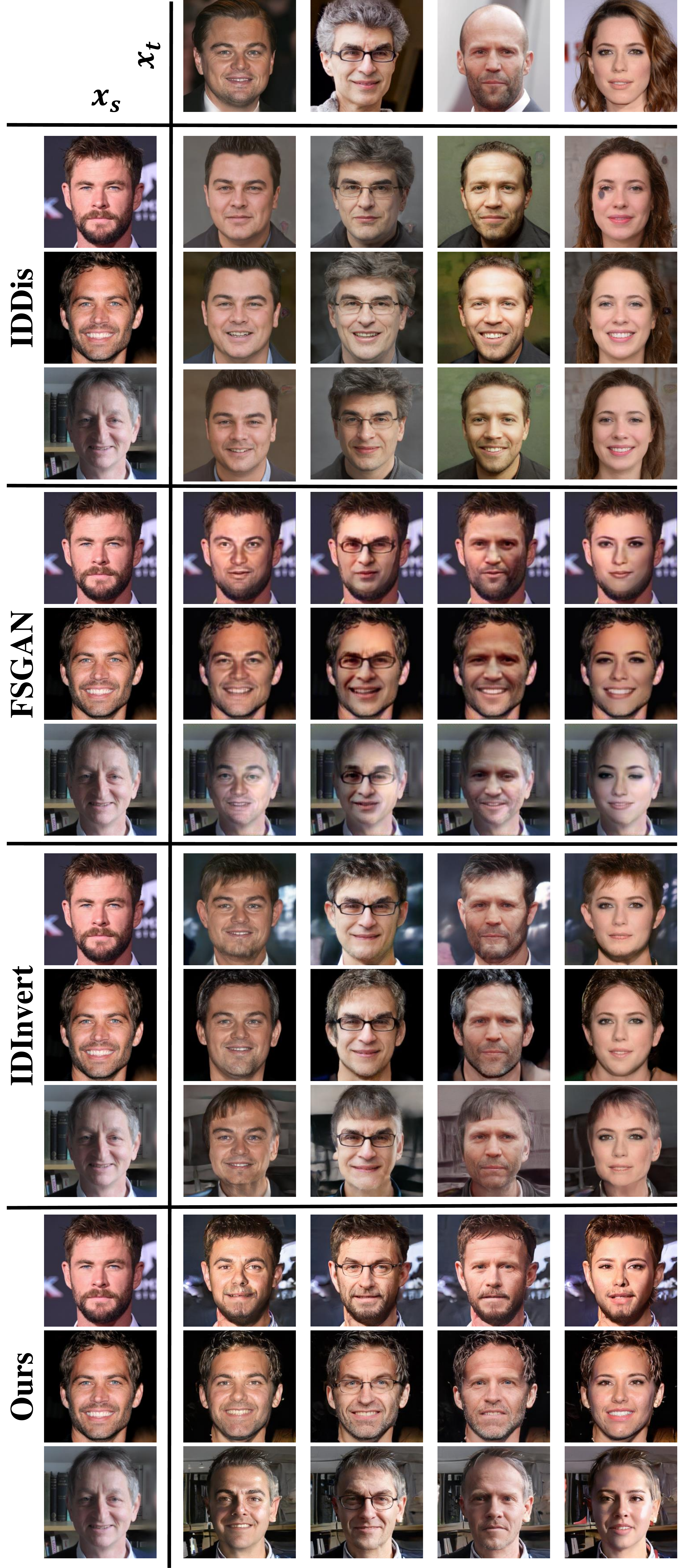}
    \caption{\textbf{Results of identity swapping.} The results of identity swapping using the images from IDInvert~\cite{id_invert}. It compares the identity swapping performance of IDDis~\cite{acm}, FSGAN~\cite{fsgan}, IDInvert~\cite{id_invert}, and our model, by inserting the identity of the target image ($x_t$) to the source image ($x_t$). }
    \label{fig:appendix_id_swapping}
\end{figure}

Due to the limited space of the main manuscripts, we attach additional details on the implementation, experimental results, and discussions in this supplementary material. 
First, Section~\ref{appendix:impl_detail} describes the details on the training settings, such as the tools and hyperparameters. 
Then, Section~\ref{sec:appendix_layer} analyzes the experimental results conducted to observe the role of each layer for the layer-wise control of attributes for de-identification. 
Next, Section~\ref{sec:contralabel attribute} provides additional examples of the controllable and uncontrollable attributes during the de-identification process. 
Lastly, the comparative analysis on our model and the existing methods for reconstruction, identity swapping, and manifold k-same algorithm are provided in Section~\ref{sec:reconstruction}, Section~\ref{sec:appendix_id_swapping}, and Section~\ref{sec:k-same}, respectively.

\section{Implementation Details}~\label{appendix:impl_detail}
We resize the input images into $256 {\times} 256$, then the generated output images have the same size. For training, we use a single NVIDIA RTX 8000 with the batch size of $16$ and $400,000$ iterations. The encoder, generator, discriminator networks are trained by Adam optimizer~\cite{adam} with the learning rate of $0.002$. Also, the weights of loss are set to $\lambda_{1}\,{=}\,1$,
$\lambda_2\,{=}\,0.01$, and $\lambda_3\,{=}\,0.05$, 
and for the manifold $k$-same algorithm, we set $k{=}3,000$.

\section{Layer-wise Control of Attributes}~\label{sec:appendix_layer}
In de-identification, the middle layers play a significant role to obscure the identity, as shown in Figure~\ref{fig:layer_w}. Out of the three types of layers, the middle layers effectively manage the attributes of glasses, mustache, and age, while the low layer only manages smiling. The hair color and style are most likely preserved unless all layers are modified. All faces in each layer are effectively de-identified even when the attributes are controlled layer-wise. For the highest degree of attribute preservation, it is most effective to modify the identity attributes in the low layers.

\section{Controllable Attributes}~\label{sec:contralabel attribute}
Figure~\ref{fig:controllable_att} illustrates the controllable and uncontrollable attributes out of 40 different attributes of CelebA-HQ~\cite{celebahq}.
Effectively disentangling the identity and non-identity attributes, FICGAN can distinguish which attribute is considered related to identity by the face recognition model~\cite{sphereface}. If an attribute to mix is considered related to identity by the face recognition model, it is reflected in the face after the mixing process; however, if an attribute to mix is considered unrelated to identity, the face recognition model excludes it in the mixing process and thus, the final result does not reflect it. As shown in the figure, most attributes are controllable, but those regarding hair and accessories, such as age-related hairstyle, hat and necklaces, are unrelated to identity and thus uncontrollable by the model. However, certain attributes, such as smiling, wearing earrings, and lipstick, seem unrelated to identity but are controllable by the model, and vice versa. This indicates an area for improvement for the face recognition model.

\section{Reconstruction}~\label{sec:reconstruction}
% Please add the following required packages to your document preamble:
% \usepackage{multirow}
\begin{table}[]
\centering
\scriptsize
% \vspace{-1.0em}
% \resizebox{1.00\linewidth}{!}{%
  \begin{tabular}{cccc}
  \hline
  Model & $\downarrow$ MSE & $\downarrow$ LPIPS & $\uparrow$ Similarity \\ \hline
  ALAE~\cite{alae} & 0.15 & 0.32 & 0.06 \\
  IDInvert~\cite{id_invert} & 0.06 & 0.22 & 0.18 \\
  pSp~\cite{psp}  & 0.04 & 0.19 & 0.58 \\
  FICGAN  & 0.02 & 0.15 & 0.30  \\ 
  \hline
  \end{tabular}
% }
\caption{\textbf{Reconstruction Performance.} }
\label{tab:recon_result}
% \vspace{-0.3cm}
\end{table}

Table~\ref{tab:recon_result} indicates the reconstructing ability of ALAE~\cite{alae}, IDInvert~\cite{id_invert}, pSp~\cite{psp}, and our model. The results indicate that our model achieves the highest scores in Mean Squared Error (MSE) and Learned Perceptual Image Patch Similarity (LPIPS)~\cite{lpips}. Since the negative loss($L_{neg}$) used for training of our model is to obscure identity from the source image, our model results in a lower similarity score despite its higher scores in MSE and LPIPS compared to pSp~\cite{psp}. Figure~\ref{fig:recon} illustrates the models' reconstructed images based on those used in pSp~\cite{psp}, indicating our model's reconstructing ability compared to others.

\section{Identity Swapping}~\label{sec:appendix_id_swapping}
\begin{figure*}[t]
\centering
        \includegraphics[width=0.99\linewidth]{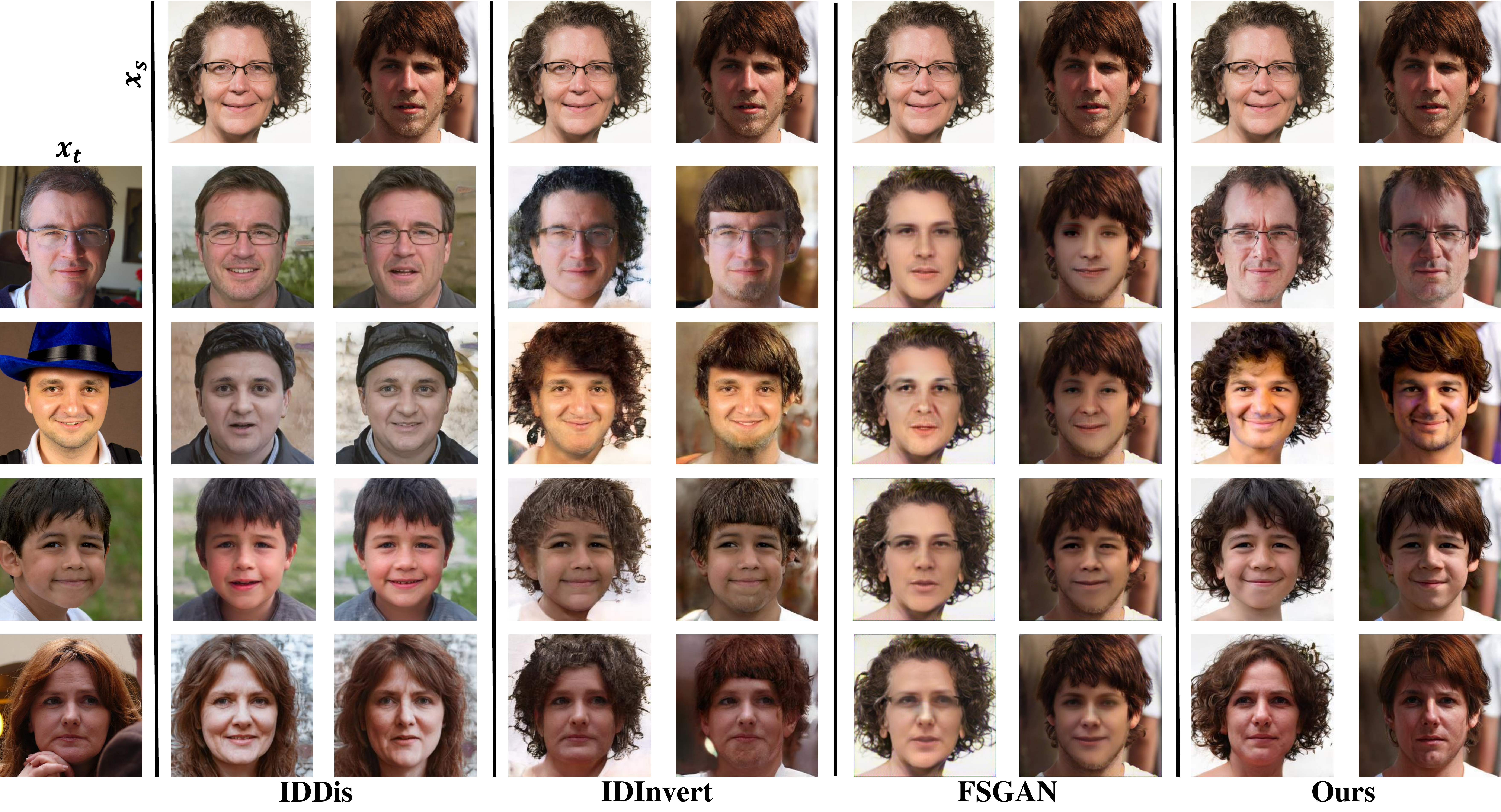}
\caption{\textbf{Additional results of identity Swapping.} The results of identity swapping using the examples from FFHQ~\cite{stylegan} unseen during the training phase. It compares the performance of IDDis~\cite{acm}, FSGAN~\cite{fsgan}, IDInvert~\cite{id_invert} and our model, by inserting the identity of the target image ($x_t$) to the source image ($x_t$).} 
        
    \label{fig:appendix_id_swapping2}
\end{figure*}

More examples of identity swapping are provided using various models and examples. 
Figure~\ref{fig:appendix_id_swapping} shows various models' results of identity swapping to compare the performance of ~\cite{acm}, ~\cite{fsgan}, ~\cite{id_invert}, and our model, using the examples from IDInvert~\cite{id_invert}.  
Figure~\ref{fig:appendix_id_swapping2} shows the results of identity swapping using the examples from FFHQ~\cite{stylegan} unseen during the training phase. The results of our model are of the highest quality among others, as shown in the included illumination, shadows, and attributes from the source images. Our model precisely reflects the direction and size of shadows, while others fail to do so. % Others fail to reflect the shadows from the source images, 
While IDDis~\cite{acm} disentangles identity and modifies hat, our model considers hat as unrelated to identity and thus excludes it from the result.

Since FSGAN~\cite{fsgan} fuses the face area of the target image with that of the source image for de-identification, it leaves the other areas in the image untouched; 
however, as shown in the first sample on the left, the eye area is found distorted due to the frame of the glasses located outside of the face area. IDInvert~\cite{id_invert} diffuses the face of the source image to the target image for effective de-identification, but other attributes, such as hair, skin tone, and the background are not preserved well. Compared to other models, our model achieves a high-quality face de-identification by effectively removing the source identity while preserving the non-identity attributes.
\section{Manifold k-same Algorithm}~\label{sec:k-same} 
\begin{figure*}[t]
\centering
        \includegraphics[width=0.99\linewidth]{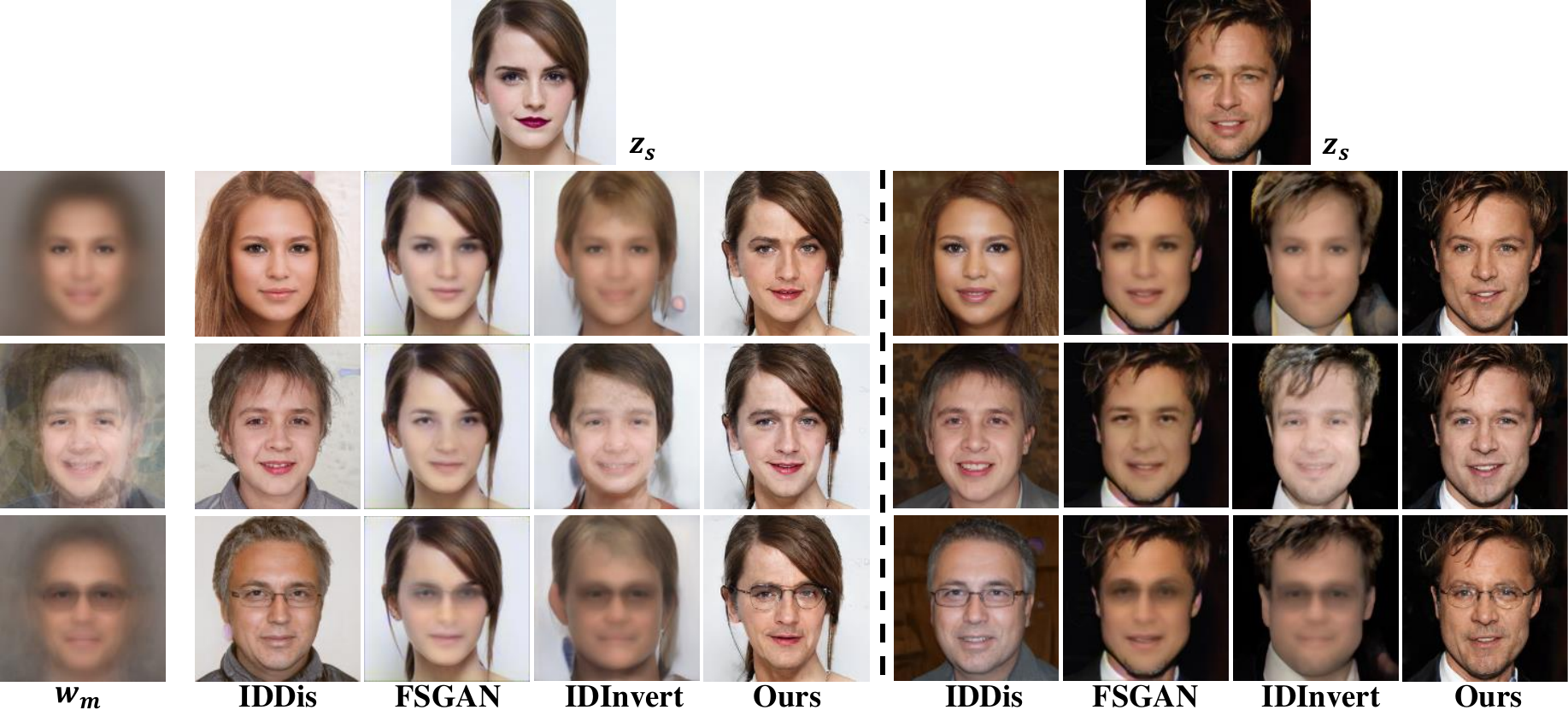}
\caption{\textbf{Additional results of attribute control with $k$-anonymity.} %Control with De-identification Image with $k$-anonymity.
The results of attribute control by incorporating $k$ number of images with the same type of attributes to IDDis~\cite{acm}, FSGAN~\cite{fsgan}, IDInvert~\cite{id_invert} and our model.
}
    \label{fig:appendix_k_mean_control}
\end{figure*}

To compare the models' performance with the manifold k-same algorithm as experimented in Figure \textcolor{red}{5}, we conduct additional experiments by employing the average faces of male, male wearing glasses, and smiling people with pale skin as $w_m$. The results of IDDis~\cite{acm}, FSGAN~\cite{fsgan}, IDInvert~\cite{id_invert}, and our model with $k$-anonymity are shown in Figure~\ref{fig:appendix_k_mean_control}. The results of FSGAN~\cite{fsgan} and IDInvert~\cite{id_invert} are blurry, while those of IDDis~\cite{acm} and our model are sharp and well reflecting the attributes $w_m$. Compared to IDInvert~\cite{id_invert} even modifying the hair style, jaw line, and background, our model preserves the non-identity attributes of $z_s$ while effectively reflecting the identity of $w_m$, which indicates superior controllability of our model. 

\section{Failure Cases}
Our method still faces some challenges in certain cases.
We observe that our model has difficulties with extreme poses or occlusions (mask, sunglasses, and unseen graphic effects like texts) that are not available in the training data.
A more fundamental source of these challenges would be due to the encoder, where it may not be able to accurately localize the facial features in such cases.
Compared to the previous methods using separate landmark detection, our method is already more robust with no accumulation error.
Our method may be further improved if such extreme data is available.
We leave this for our future research direction.
\begin{figure}[t]
\centering
        \includegraphics[width=0.95\linewidth]{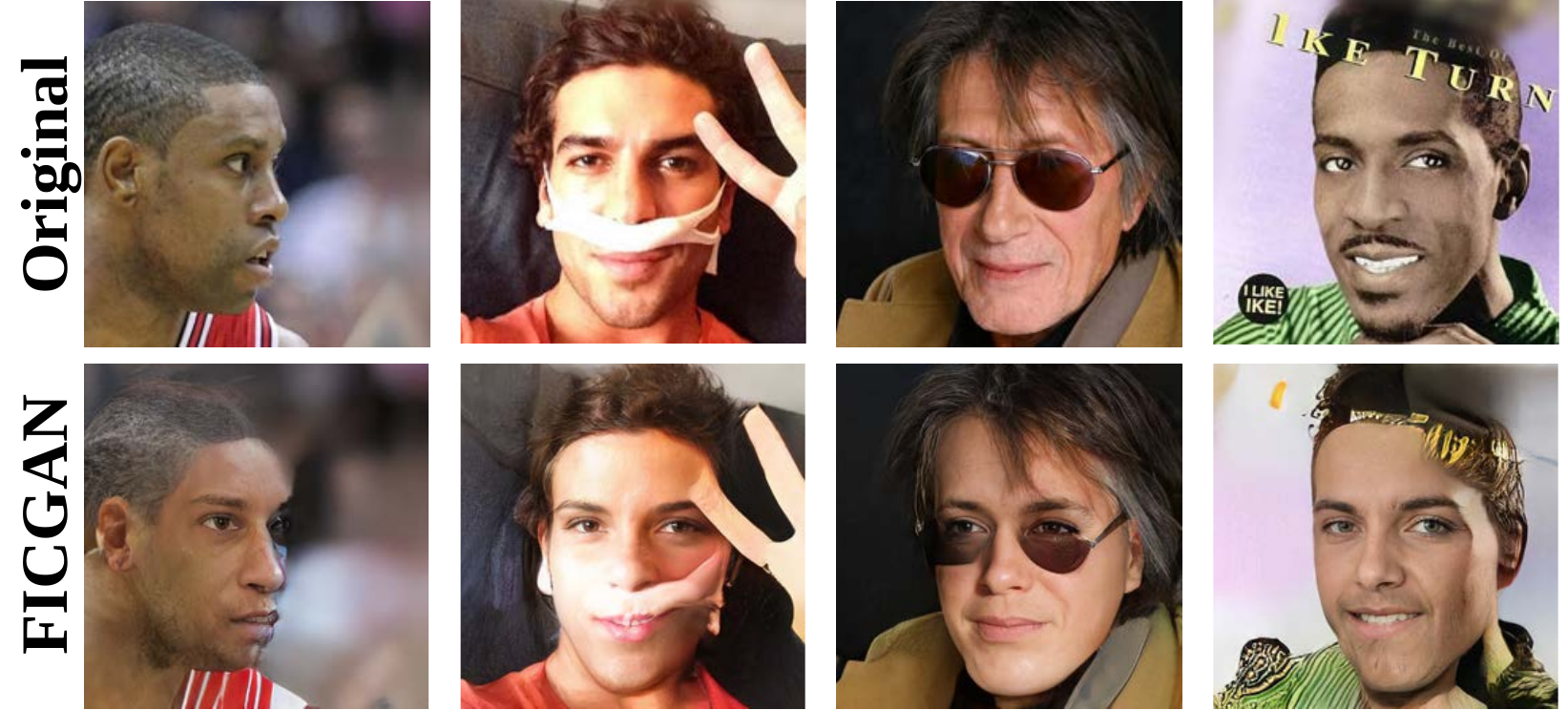}
\caption{\textbf{Challenging cases for FICGAN.}
        }
    \vspace{-0.3cm}
    \label{fig:challenges_case} 
    \vspace{-0.1cm}
    % \vspace{-1.5em}
\end{figure}

\end{document}